\documentclass[10pt,twocolumn,letterpaper]{article}

\usepackage{cvpr}
\usepackage{times}
\usepackage{epsfig}
\usepackage{graphicx}
\usepackage{amsmath}
\usepackage{amssymb}
\usepackage{textcomp} 
\usepackage{mathtools}
\usepackage{gensymb} 
\usepackage{xfrac} 
\everymath{\displaystyle} 
\usepackage{siunitx} 
\usepackage{booktabs}
\usepackage{tabularx}
\usepackage{caption} 
\usepackage{diagbox}
\usepackage[hyphens,spaces,obeyspaces]{url} 

\newcommand\HR{\mathit{HR}} 
\newcommand\RR{\mathit{RR}} 

\usepackage[breaklinks=true,bookmarks=false]{hyperref}
\usepackage[square,comma,numbers,sort&compress,sectionbib]{natbib}

\cvprfinalcopy 

\def\cvprPaperID{15} 

\ifcvprfinal\fi
\begin{document}

\title{Remote Photoplethysmography: Rarely Considered Factors}

\author{Yuriy Mironenko \hspace{35pt} Konstantin Kalinin \hspace{35pt} Mikhail Kopeliovich \hspace{35pt} Mikhail Petrushan\\
Southern Federal University, Center of Neurotechnologies\\
Rostov-on-Don, Russian Federation\\
{\tt\small kop@km.ru}
}


\maketitle

\begin{abstract}

Remote Photoplethysmography (rPPG) is a fast-growing technique of vital sign estimation by analyzing video of a person. Several major phenomena affecting rPPG signals have been studied (e.g. video compression, distance from person to camera, skin tone, head motions). However, to develop a highly accurate rPPG method, new, minor, factors should be investigated. First considered factor is irregular frame rate of video recordings. Despite of PPG signal transformation by frame rate irregularity, no significant distortion of PPG signal spectra was found in the experiments.  Second factor is rolling shutter effect which generates tiny phase shift of the same PPG signal in different parts of the frame caused by progressive scanning. In particular conditions effect of this artifact could be of the same order of magnitude as physiologically caused phase shifts. Third factor is a size of temporal windows, which could significantly influence the estimated error of vital sign evaluation. It follows that one should account difference in size of processing windows when comparing rPPG methods. Short series of experiments were conducted to estimate importance of these phenomena and to determine necessity of their further comprehensive study.

\end{abstract}

\section{Introduction}

Photoplethysmography (PPG) is an estimation of blood volume changes in tissue by measuring characteristics of light either passed through tissue (mostly by contact PPG) or reflected from tissue (mostly by remote PPG, also referred as rPPG). rPPG signal which carries information about blood volume changes caused by heart beats. Typically, two sources of rPPG signal are considered. First is optical absorption by hemoglobin molecules~\cite{Prahl2020}. Early theory assumed that changes of optical density of tissue are produced by pulse wave passing through blood vessels~\cite{Teplov2014}. Alternative explanation are elastic deformations of capillary bed by pulse wave in underlying arteries~\cite{Kamshilin2015}. Second source of rPPG signal is slight motion of body or head caused by pulse wave in arteries. Particularly, head makes tiny tilt movements due to carotid pulsation. Generally, both sources relate mostly to arteries pulsation, which generates capillaries motion in normal direction and head motion in tangential direction to skin surface.

rPPG methods evaluate vital signs. Thus, proper accounting of factors affecting this evaluation and its interpretation is critical. A variety of rPPG algorithms have been developed since early 2000s~\cite{Verkruysse2008, Wu2000, Wu2003, Wieringa2005} and significantly improved over the last years: some are based on mathematical models involving physical properties of light reflectance such as CHROM~\cite{Haan2013} and POS~\cite{Wang2017_2}, others are based on deep learning approaches, such as~\cite{Chen2018, Yu2019RecoveringRP, Sabokrou2020}. Recent methods demonstrate promising accuracy of HR estimation, reaching mean absolute error (MAE) lower than 3 beats per minute (bpm) on particular datasets~\cite{Blackford2017, Chen2018, Finzgar2018}. Such low error is reached by optimization of major factors such as processing pipeline or deep learning architecture, robust tracking of areas for rPPG signal retrieving. To further improve rPPG methods, additional factors which have minor effect on accuracy should be accounted. Another motivation for considering these new factors is the importance - due to narrower dynamic range - of high accuracy for retrieving of differential characteristics such as short-term HR variability or PPG phase shift. Next important concern is reasonable choice of better rPPG method for particular conditions. Particularly, the question is: is it correct to directly compare the rPPG methods that were trained on video sequences of various durations? For example, some method produces HR estimates on time series of 4~sec duration, another -– on 20~sec series. They are both tested on a dataset where ground truth HR values were computed on ECG signal within 10~sec duration temporal windows. If particular method performs better, it means that it provides HR estimations closer to ground truth values calculated in the said manner. Further, if ground truth~(GT) values are recalculated for another duration of temporal window (4~sec or 20~sec), another method could become better, meaning it provides estimates closer to new GT values. In variety of works where rPPG methods are compared, such information about duration of temporal windows for computing GT values and estimation of HR values is omitted~(see Table~\ref{table:metComp}). In our opinion, such omitting of this information could lead to misunderstanding of optimal conditions where comparing methods perform better.

The following attributes of PPG signal are typically estimated: heart rate~\cite{Verkruysse2008, Poh2010, Wu2000}, heart rate variability~\cite{WangBrinker2018} respiratory rate~\cite{Kooij2019}, phase shift of PPG signal on different skin areas~\cite{Verkruysse2008}. These attributes depend on health state and functional state of a subject; thus, they could be used as indicators of such states. In particular, heart rate is used for physical load assessment~\cite{Jindal2016}, while heart rate variability is for stress estimation~\cite{Olsson2010, Sun2012, Mcintyre2016}. Regarding PPG signals phase shift, initially it was interpreted as a correlate of abnormal health state, particularly – migraine~\cite{Kamshilin2018_2}. Later, contradictory explanation was introduced~\cite{Kamshilin2015}: opposite phases of PPG signal in neighboring areas are caused by elastic nature of skin in the following manner. Pulse wave in an artery deforms capillary bed to skin surface  above an artery. Due to skin elasticity, neighboring areas can move in opposite direction. Authors gave intuitive explanation by considering a sponge: when some part of it is moving towards under pressure, neighboring areas can move in opposite direction. Such different interpretation of the same phenomenon led to different conclusions on person\textquotesingle s state. This example demonstrates importance of correct accounting of rPPG signal nature for proper inferring about person\textquotesingle s state. 

Another factor which should be properly considered in rPPG method applications is a process of image acquisition by a camera. Properties and artifacts of such acquisition affect rPPG signal and can lead to incorrect conclusions. PPG signal can be extracted and enhanced in video captured even by common webcams~\cite{Rustand2012, Haan2013, Feng2015a}. rPPG signal in such video is characterized by relatively low signal to noise ratio (SNR). Therefore, considering and suppression of possible artifacts is critical for SNR increasing.

Typically, processing of color signals by classic signal processing methods (which are not based on deep learning models) is based on assumption of regular inter-frame duration. Particularly, computing Fourier spectra (or wavelet transform) doesn\textquotesingle t account information about irregular inter-frame duration. However, some rPPG datasets (like VIPL-HR~\cite{Niu2018_1} and PURE~\cite{Stricker2014}) has irregular frame rate in their videos, and \cite{Fletcher2015} claims importance of irregular frame rate for the correct HR estimation.

One more video-related factor is a rolling shutter effect, also known as progressive scanning of a frame. Video frame is not obtained in a single moment in time, since acquisition of each frame is a continuous process. As a result, one part of a frame is captured earlier than another, causing appearance of phase shift artifacts. Due to phase shift can be treated as diagnostic factor~\cite{Teplov2014}, it\textquotesingle s important to exclude potential artifact of progressive scanning in phase shift of rPPG signals. We didn\textquotesingle t considered rolling shutter as a source of HR estimation errors, however such phenomena as Pulse Transit Time may be affected if artifact is strong enough.

Next factor -- camera\textquotesingle s anti-banding -- is not studied in our work. However, it possibly can affect rPPG method and, thus, could merit further research. Anti-banding filter suppresses flicker taking place in video recordings under lamplight conditions. The flicker is produced by beat of frame capturing frequency and artificial light source frequency. Due to frequency of this beat can be close to HR frequency, anti-banding could suppress rPPG signal.

In this work, we review factors affecting accuracy of HR estimation focusing on video capture-related aspects, and proper comparing of rPPG methods. Namely, we consider irregular frame rate and rolling shutter effect. In this work we rather aim at illustrating possible effect of this factors on HR evaluation on several examples, than providing comprehensive statistical evaluation of such effect. Regarding correct comparing of rPPG methods, we focus on temporal window selection for HR computing both in rPPG processing and in ground truth labels.

\section{Related works}

In this section, we review available works in two areas: 1~-- rPPG methods standardization and matching and 2~-- studies of video capture-related artifacts and their suppression.

\begin{table}
\centering
\begin{tabularx}{0.47\textwidth}{p{2.55cm} | >{\raggedright\arraybackslash}X}
\toprule
\textbf{Paper or}\par\textbf{proposed method}    & \textbf{Methods compared with} \\
\midrule
CHROM~\cite{Haan2013} & ICA~\cite{Poh2011}, PCA~\cite{Lewandowska2011}
\\ \cline{2-2}
SAMC~\cite{Tulyakov2016} & ICA~\cite{Poh2011}, BCG~\cite{Balakrishnan2013}, Li2014~\cite{Li2014}, CHROM~\cite{Haan2013}
\\ \cline{2-2}
Huang2016~\cite{Huang2016} &  CHROM~\cite{Haan2013}, ICA~\cite{Poh2011}
\\ \cline{2-2}
POS~\cite{Wang2017_2} & G~\cite{Verkruysse2008}, PCA~\cite{Lewandowska2011}. ICA~\cite{Poh2011},  CHROM~\cite{Haan2013}, PBV~\cite{Haan2014}, 2SR~\cite{Wang2016}
\\ \cline{2-2}
Coppetti2017~\cite{Coppetti2017} & Four iOS applications for contact and non-contact PPG
\\ \cline{2-2}
cICA~\cite{Macwan2018} & cICA~\cite{Macwan2018}, ICA~\cite{Poh2011}, PCA~\cite{Lewandowska2011}, G~\cite{Verkruysse2008}, CHROM~\cite{Haan2013}, POS~\cite{Wang2017_2}
\\ \cline{2-2}
SparsePPG~\cite{Nowara2018} & DistancePPG~\cite{Kumar2015}, CHROM~\cite{Haan2013}, ICA~\cite{Poh2011}
\\ \cline{2-2}
HR-CNN~\cite{Spetlik2018} & 2SR~\cite{Wang2016}, Li2014~\cite{Li2014}, SAMC~\cite{Tulyakov2016}
\\ \cline{2-2}
SynRhythm~\cite{Niu2018_2} & ICA~\cite{Poh2011}, BCG~\cite{Balakrishnan2013}, Li2014~\cite{Li2014}, CHROM~\cite{Haan2013}, Niu2017~\cite{Niu2017}, SAMC~\cite{Tulyakov2016}, Hsu2014~\cite{Hsu2014}
\\ \cline{2-2}
DeepPhys~\cite{Chen2018} & Estepp2014~\cite{Estepp2014}, McDuff2014~\cite{McDuff2014}, BCG~\cite{Balakrishnan2013}, CHROM~\cite{Haan2013}, POS~\cite{Wang2017_2}, SAMC~\cite{Tulyakov2016}
\\ \cline{2-2}
Deep-HR~\cite{Sabokrou2020} & ICA~\cite{Poh2010, Poh2011}, BCG~\cite{Balakrishnan2013}, Li2014~\cite{Li2014}, CHROM~\cite{Haan2013}, SAMC~\cite{Tulyakov2016}, Hsu2014~\cite{Hsu2014}, RithmNet~\cite{Niu2019}, HR-CNN~\cite{Spetlik2018}
\\ \cline{2-2}
PhysNet~\cite{Yu2019RecoveringRP} & SynRhythm~\cite{Niu2018_2}, HR-CNN~\cite{Spetlik2018}, DeepPhys~\cite{Chen2018}              
\\ \cline{2-2}
RithmNet~\cite{Niu2019} & ICA~\cite{Poh2011}, BCG~\cite{Balakrishnan2013}, Li2014~\cite{Li2014}, CHROM~\cite{Haan2013}, SAMC~\cite{Tulyakov2016}
\\ \cline{2-2}
STVEN +\par rPPGNet~\cite{yu2019_2} & ICA~\cite{Poh2011},  CHROM~\cite{Haan2013}, Li2014~\cite{Li2014}, SAMC~\cite{Tulyakov2016}, SynRhythm~\cite{Niu2018_2}, HR-CNN~\cite{Spetlik2018}, DeepPhys~\cite{Chen2018} 
\\
\bottomrule
\end{tabularx}
\caption{Papers comparing methods without considering the size of the temporal window for HR calculation.}
\label{table:metComp}
\end{table}

With significant growth of amount of works in contact and contactless PPG, several reviews appeared where authors aimed to systematize different approaches and conditions of their applicability. In~\cite{antink2019} authors considered a problem of weakly standardized experimental setups and datasets for PPG, emphasizing contrast between usage of industrial-grade equipment in experimental setups and marketing rPPG as low-cost technology. Such weak standardization makes matching of HR estimation results obtained using different approaches complicated and even impossible in some cases. Papers listed in Table~\ref{table:metComp} compared rPPG methods of HR estimation regardless of their processing windows. In rare cases, authors provide optimization results of varying processing window size for a particular method~\cite{Haan2013, Tulyakov2016,  Finzgar2018, Macwan2018}. Nevertheless, existing methods are usually implemented with ``default'' settings provided by their authors and compared with no attention to their processing window size.

The following datasets are commonly used for training and evaluation of rPPG approaches: Mahnob-HCI~\cite{Soleymani2012}, MMSE-HR~\cite{Zhang2016}, PURE~\cite{Stricker2014}, and VIPL-HR~\cite{Niu2018_1}. They consist of videos with subjects been sitting in front of a camera for 10 seconds or more. The number of subjects varies from 10 (PURE) to 107 (VIPL-HR). The reference data was recorded from contact sensors as electrocardiogram or photoplethysmogram. Ground truth HR values could be calculated in different ways from reference data, if using different methods (peak detection-based or spectrum-based). These GT values calculation as well as routines of data conversions are handled by frameworks, which are reviewed below.

One way to standardize evaluation and matching of rPPG methods is to compare them in homogeneous environment with universal rules of computing evaluations and ground truth values on the same temporal windows.
To the best of our knowledge, there are a few open-source tools implementing several steps of rPPG pipeline: iPhys-Toolbox~\cite{mcduff2019}, Kooij2019~\cite{Kooij2019} and PPGI-Toolbox~\cite{pilz2019}. The frameworks provide several methods of rPPG signal processing, such as, ICA~\cite{Poh2010}, CHROM~\cite{Haan2013}, POS~\cite{Wang2017_2}, BCG~\cite{Balakrishnan2013}, 2SR~\cite{Wang2016}, LGI~\cite{pilz2019}, as well as some classical approaches of image processing and HR estimation using periodogram. However, existing tools are mostly available for Matlab, while another research environments (e.g. Python-based) are not well covered. There is still a lack of general, readily available, open-source rPPG framework allowing to research variety of methods (including machine-learning) and to evaluate them on public datasets.

Benchmarks and challenges is a common way to match efficiency of different approaches. While they are widely used in most of computer vision and machine learning areas, they just start appearing in rPPG. 
RePSS (Remote Physiological Signal Sensing)~\cite{Li2020} is the first challenge in the area, conducted in conjunction with CVPR 2020.
There is also another benchmark on HR evaluation, ``rPPG benchmark''~\cite{Kooij2019}. To date, it contains only a single private dataset with a single result on it from the authors\textquotesingle method.

Factors related to video capturing and video compression are intensively studied to reveal their effect on rPPG accuracy~\cite{mcduff2018, zhao2018, yu2019_2}. Video compression strongly suppresses rPPG signal in video. In~\cite{zhao2018} authors described types of artifact in rPPG signal caused by compression and proposed a framework to deal with compressed video. In particular, they found red and blue color components are mostly affected by video compression artifacts in a case of low-bitrate video and proposed single-channel processing approach (greed color component) which outperforms multichannel-based approaches on low-bitrate video. Yu developed STVEN autoencoder~\cite{yu2019_2} to convert video from one bitrate to another with aim of enhancement of rPPG signal. McDuff proposed Deep Super Resolution network~\cite{mcduff2018} for low resolution video which also allows to enhance rPPG methods on compressed video.

Mentions of phase shift could be traced up to Verkruysse\textquotesingle s work~\cite{Verkruysse2008}. It was studied recently at~\cite{moco2018_1}, and it was demonstrated that phase shift may be used to distinguish rest and exercise condition of the subject. So-called ``rolling shutter'' effect, caused by progressive scan presumably on CMOS sensors, is a well-known source of artifacts, efforts was made to measure~\cite{Oth2013} and compensate~\cite{Lao2018} the effect. No prior works were found which evaluate possibility of rPPG signal could be affected by rolling shutter. However, in~\cite{Verkruysse2008} authors noticed the possible influence of automatic gain correction on the phase shift and found its influence insignificant.

In~\cite{Fletcher2015}, authors claimed significant effect of frame rate irregularity on HR estimations, however, they didn\textquotesingle t separate ``additional background processes'' from ``frame rate jitter''. Other than that, irregular fps is poorly studied in rPPG. 

\section{Rolling Shutter Experiment}
\begin{figure}
	\centering
	\captionsetup{justification=centering}
	\includegraphics[width=80mm]{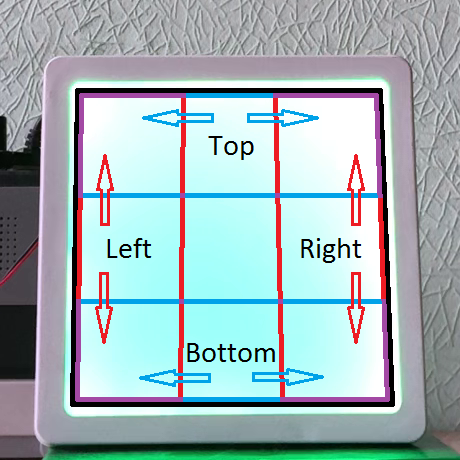}
	\caption{The lamp divided into four regions that were used in the rolling shutter experiment. Best viewed in color.}
	\label{fig:LEDsetup}
\end{figure}
\begin{figure}
	\centering
	\captionsetup{justification=centering}
	\includegraphics[width=80mm]{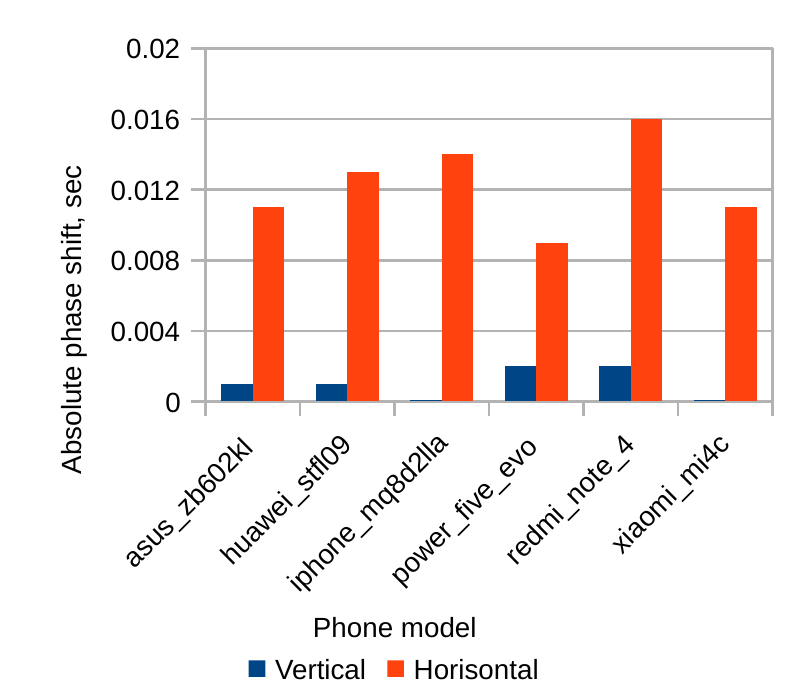}
	\caption{Median values of estimated absolute amplitude shift.}
	\label{fig:phase}
\end{figure}
To evaluate the impact of rolling shutter effect on the retrieving of the phase shift over the picture, several PPG records from MMSE-HR dataset~\cite{Zhang2016} were used. The records were played back through the model object -- a big LED lamp with finely-controllable brightness, designed to exclude any phase shift along its surface (Figure~\ref{fig:LEDsetup}).
Six smartphones (Figure~\ref{fig:phase}) were simultaneously used to capture video from the model object. If there is no rolling shutter effect, there should be no phase shift in the captured videos. Otherwise, observable phase shift is caused by the rolling shutter.

To verify the existence of a phase shift, the four brightness signals were extracted from a video as time series of averaged grayscale intensities over four regions: from top, bottom, left, and right regions of the model object (see Figure~\ref{fig:LEDsetup}). Then the phase shift is calculated along the vertical axis (i.e. between top and bottom signals) and horizontal axis (i.e. between left and right signals). Rolling shutter effect expected to produce phase shift along the progressive scan direction, while phase shift along orthogonal direction is expected to be close to zero.

\subsection{Experimental Hardware}

The experiment required light source (``modulator'') that can be modulated using digital signal at a relatively high sample rate.
The hardware pipeline shouldn\textquotesingle t have its own significant influence at a frequencies that are typical for heart rate and its subproducts in a form of skin color changes. Thus, the modulator should support low frequency modulation (even constant value).

\subsubsection{Computer Connectivity}
To simplify the task of digital-to-analogue and vice-versa conversion and to use an already existing well-tried solution, it was considered to utilize a sound card as a fast ADC/DAC (Analog-to-Digital Converter / Digital-to-Analog Converter) module and GNU Radio~\cite{GnuRadio} as a backend.
A hardware customization of a sound card was required. A regular sound card have DC (Direct Current) decoupler capacitors both at line input and line output. It is reasonable filtering procedure for an average sound signal but not for the photoplethysmography. Genius Sound Maker 5.1 PCI sound card was used having the capacitors well recognizable. The capacitors were bypassed and after that modification the "zero" digital reading was strictly attached to the ADC/DAC idle voltage of 2.5 volts DC.
The sound card sample rate is 44100 per second that is enough for the experimental purposes.

\subsubsection{The Modulator}
The modulator task is to produce a light at a given sample rate. The common solutions can lead to an incorrect experiment setup. A PWM (Pulse Width Modulation) that is widely used to dim the power of LED does not suit our task as it can produce moire effect. White lamps utilise phosphor LEDs that have an afterglow and a wide spectral characteristic. So let\textquotesingle s formulate the requirements for the modulated light source:
\begin{itemize}
\item do not use a PWM modulation to achieve the required light output;
\item do not have an afterglow;
\item generated light should be perceptible by the camera.
\end{itemize} 
A LED lamp light with a diffuser was customized. White LED strip was replaced with a green LED strip. Green LED produce direct light emission without phosphor thus having no afterglow. Another reason is that LED wavelength is in the middle of the visible range of a generic camera.
A linear amplifier was developed to drive the LED strip within sound card DAC in a full range of possible input. After that the GNU Radio backend can use sound card linear output to drive the Modulator light.

\subsection{Results}
\subsubsection{Phase Shifts caused by Rolling Shutter effect}
We was able to identify phase shift of up to 0.02 sec. for all the smartphones used. It was always registered along horizontal axis, and vertical axis has order-of-magnitude smaller phase shift (Figure~\ref{fig:phase}). This difference exists for all of the PPG signal used.

For comparison, phase shift of 2 frames was demonstrated in~\cite{Verkruysse2008}, ``\textit{0.067s, or 34 degrees at a HR of 1.43~Hz}''. Phase shift of 0.02 sec, caused by progressive scan, is about 10 degrees at HR of 1.43~Hz (mind that this ``\textit{progressive scan shift}'' is not related to the actual heart rate, and it's natural to measure it in seconds, not in degrees). 

Moço~\etal~\cite{moco2018_1} quote 14.8$\degree$ of cheek-forehead phase shift at the period of rest and 17.5$\degree$ at the period of exercise. HR is not reported unfortunately, but 0.02~sec is 7.2$\degree$ at 60~bpm and 9.6$\degree$ at 80~bpm.

So we conclude that rPPG phase shift caused by the progressive scan is big enough to produce significant disturbance to the phase-related experiments, especially the ones where \textit{difference} of phase shifts is estimated. Special attention should be paid to avoid it or filter it out. Simplest possible approach is to take sensor orientation into account and demonstrate effect after the sensor is rotated to 90 deg.

\subsubsection{Phase Shifts oscillation}
More important is that 0.02 sec is not the constant shift. Actually, phase shift slowly oscillates from effectively zero up to maximum and back to zero. These oscillations don\textquotesingle t look like random, but seem to be quite periodic. These oscillations happen for all the six smartphones used in our experiments.

We actually have no any obvious explanation for this oscillations -- either by video capturing artifact or by experiment setup flaw. However, this effect may be quite important because of it\textquotesingle s slowness - because period may be as long as tens of seconds, it\textquotesingle s possible to get serious artifacts just because one may accidentally catch close-to-zero part of the curve in one experiment, and close-to-maximum part in another, producing false positive recognition of phase shift because of it.

For example, in ~\cite{moco2018_1} phase shift difference between rest and exercise periods is 2.7$\degree$. With progressive scan phase shift oscillating from 0$\degree$ to 7.2$\degree$, it's easy to get a false positive result. Moreover, as long as progressive scan phase shift is never negative over this oscillations, multiple experiments may not compensate it.

\section{Irregular Frame Rate Experiment}
We captured 10 different PPG signals from each of 6 different smartphones. This resulted in 60 video records, which were split into 279 10-seconds pieces -- we need small pieces to avoid potential artifacts to be averaged out.

Video stream encoded with average codec actually contain ``presentation time'' for each frame - the moment when a frame should be shown. In case of regular frame rate these values follow at regular intervals, otherwise not. This information is used by decoder to render the video.
FFprobe utility was used to extract the frame ``presentation times''. We consider naming the ``presentation time'' as ``frame timestamp''.
To understand if irregular frame rate affects HR estimations, we used essentially the same experimental setup as described above, but with different processing of the resulting videos. 
Specifically, we converted every video to the ``timestamped'' irregular signal - i.e. average intensity of the whole model object was calculated for every frame and then stored along with the timestamp of this frame. This "timestamped" irregular signal was then interpolated to the regular 44100 Hz signal using two methods:
\begin{itemize}
    \item ``good'' one, which uses actual timestamps from the video in the process of interpolation;
    \item ``bad'' one, which uses wrong timestamps, calculated as if frame rate of the video is constant (i.e. taking timestamp of the first frame, timestamp of the last frame, and dividing period between this timestamps equally to all the frames between).
\end{itemize}

``Bad'' signal, if compared to the ``good'' one, if somewhat ``squeezed'' in some places and somewhat ``stretched'' in other places, as expected. To understand if this squeezes-and-stretches are big enough to cause any significant disturbance, we evaluated:
\begin{itemize}
    \item difference between ``good'' and ``bad'' signals;
    \item difference between their spectra.
\end{itemize}
\subsection{Results}
\begin{figure*}
	\centering
	\captionsetup{justification=centering}
	\includegraphics[width=170mm]{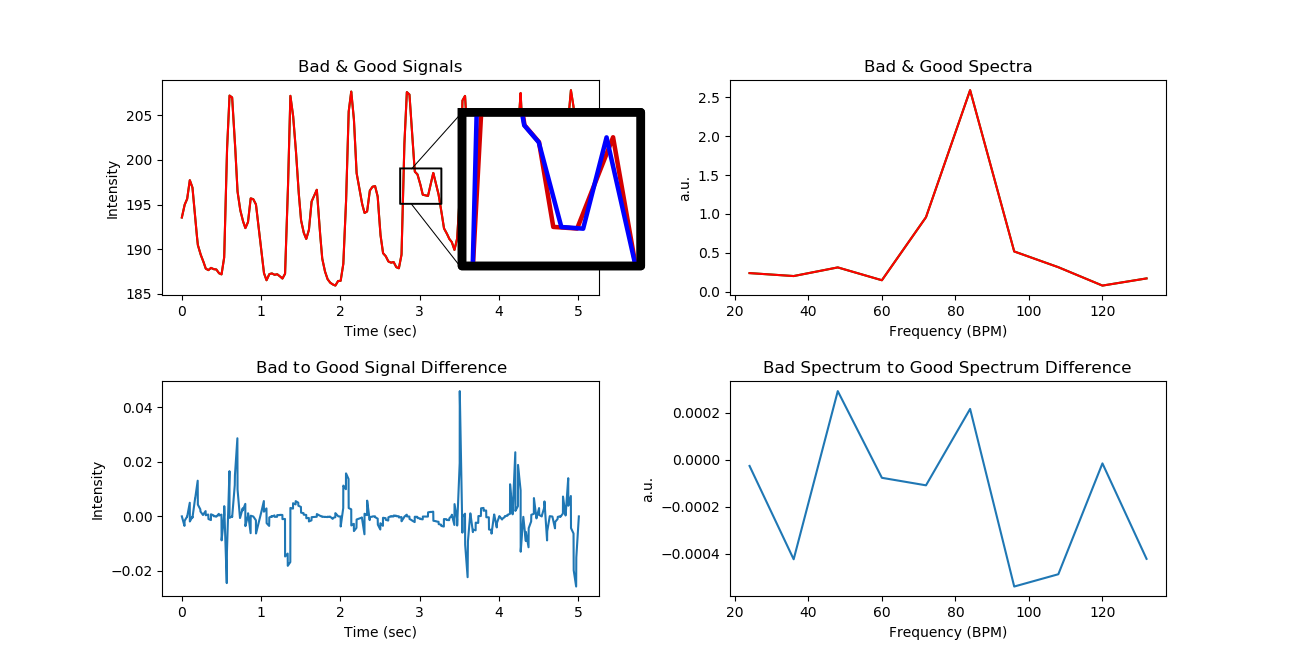}
	\caption{Comparison of timestamp-aware (good, colored in blue in upper figure) and timestamp-ignorant (bad, colored in red in upper figure) signals and their power spectra. Due to insignificant signals and spectra difference "bad" and "good" plots looks same on this scale. Displacements on zoomed view is not to scale.}
	\label{fig:irregular}
\end{figure*}
All 6 smartphones produce videos with irregular frame rate. But our results contradict to~\cite{Fletcher2015} -- difference caused by irregular frame rate is mostly inessential in our experiments. Both amplitude and spectral differences of ``good'' and ``bad'' signals are orders of magnitude smaller than amplitude of the signal and absolute spectrum~(Figure~\ref{fig:irregular}).

This result remains the same for the different cameras with codecs with different irregularity structures.
\section{Effect of temporal window on HR estimation}
\subsection{Methods}
The effect of using different sizes of temporal window was investigated on a subset of Fantasia Database~\cite{Iyengar1996}, from PhysioNet~\cite{Goldberger2000}. The subset contains 120-minute records of ten healthy subjects: five young (21--34 years old) and five elderly (68--85 years old). Subjects had been lay in resting state for 120 minutes watching a movie while electrocardiographic (ECG) signals were collected. The ECG frame rate was 250~Hz. Each heartbeat was annotated using an automated arrhythmia detection algorithm~\cite{Iyengar1996}, and each beat annotation was verified by visual inspection. The interbeat (RR) interval time series for each subject were then computed.

Let $s \in \{0,5,\dotsc,60\}$ sec be the size of a temporal window $W^s$ which contains $N$ annotated heart beats. ``0-sec'' window size denotes a single $\RR$ interval. Then, the HR value was calculated as an average of RR intervals $\RR_i$ calculated within the $W^s$:
\begin{equation}
\label{HR_W}
\HR\left(W^s\right)=
\begin{cases}
    \frac{1}{\RR},                       & \text{for } s = 0 \text{ sec} \\[0.2cm]
    \frac{N-1}{\sum_{i=1}^{N-1}{\RR_i}}, & \text{for } 5 \leq s \leq 60 \text{ sec}
\end{cases}
.
\end{equation}

Heart rates, which are averaged within temporal windows of $s_1$ and $s_2$ sizes $\left(s_1>s_2\right)$, were compared as follows. The whole time series were divided into sequences of $s_1$-sec windows.
In each window $W^{s_1}$ another shorter sliding window $W^{s_2}_i \left(\subset W^{s_1}\right)$ of $s_2$-sec duration was selected with a step of 5 seconds, $i=0, \dotsc,\left(s_1-s_2\right)/5$. In case of $s_2=0$, all RR intervals within the $W^{s_1}$ window were selected.
Next, HR values were computed using \eqref{HR_W} for $W^{s_1}$ window and for each $W^{s_2}_i$. The differences between HR values were averaged as relative difference:
\begin{equation}
d\left(W^{s_1}, W^{s_2}_i\right)=
    \frac{\left|\HR\left(W^{s_1}\right) - \HR\left(W^{s_2}_i\right)\right|}
    {\HR\left(W^{s_1}\right)}\cdot100\%
.
\end{equation}

The larger $d$-value indicates greater difference of HR values estimated within different temporal windows. 
Implementation of this method in Python code is publicly available\footnote{\ifcvprfinal \url{https://github.com/Simplar/Effect-of-temporal-window-on-HR-estimation}\else anonymized link to source code\fi}.

\begin{figure}
    \centering
    \captionsetup{justification=centering}
	\includegraphics[width=80mm]{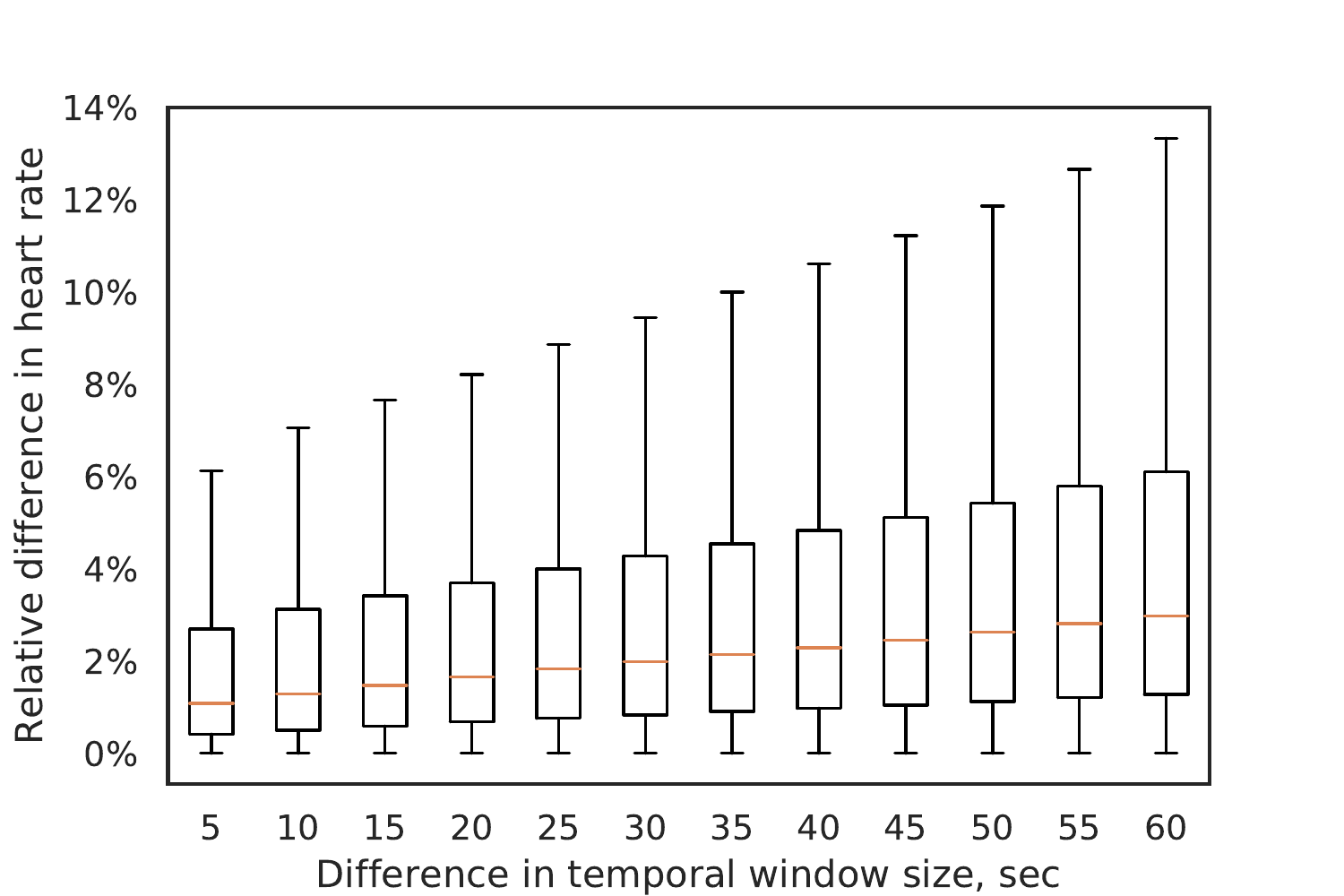}
	\caption{Box plot of distributions of differences in HR values estimated for various sizes of temporal windows.}
	\label{fig:DiffBox}
\end{figure}
\begin{table*}
\centering
\begin{tabular}{c|llllllllllll}
\toprule
\backslashbox{$s_1$}{$s_2$}
&    0  &    5  &    10 &    15 &    20 &    25 &    30 &    35 &    40 &    45 &    50 &    55 \\
\midrule
5  &  2.7\% &     - &     - &     - &     - &     - &     - &     - &     - &     - &     - &     - \\
10 &  3.3\% &  1.7\% &     - &     - &     - &     - &     - &     - &     - &     - &     - &     - \\
15 &  3.6\% &  2.1\% &  1.2\% &     - &     - &     - &     - &     - &     - &     - &     - &     - \\
20 &  3.8\% &  2.4\% &  1.5\% &  0.9\% &     - &     - &     - &     - &     - &     - &     - &     - \\
25 &  4.0\% &  2.6\% &  1.8\% &  1.2\% &  0.8\% &     - &     - &     - &     - &     - &     - &     - \\
30 &  4.2\% &  2.8\% &  2.1\% &  1.5\% &  1.1\% &  0.6\% &     - &     - &     - &     - &     - &     - \\
35 &  4.3\% &  2.9\% &  2.2\% &  1.8\% &  1.3\% &  0.9\% &  0.6\% &     - &     - &     - &     - &     - \\
40 &  4.3\% &  3.1\% &  2.4\% &  2.0\% &  1.6\% &  1.2\% &  0.8\% &  0.5\% &     - &     - &     - &     - \\
45 &  4.4\% &  3.2\% &  2.5\% &  2.1\% &  1.8\% &  1.4\% &  1.1\% &  0.7\% &  0.4\% &     - &     - &     - \\
50 &  4.5\% &  3.2\% &  2.7\% &  2.3\% &  2.0\% &  1.6\% &  1.3\% &  1.0\% &  0.7\% &  0.4\% &     - &     - \\
55 &  4.5\% &  3.3\% &  2.7\% &  2.4\% &  2.1\% &  1.8\% &  1.5\% &  1.2\% &  0.9\% &  0.6\% &  0.4\% &     - \\
60 &  4.6\% &  3.3\% &  2.8\% &  2.4\% &  2.1\% &  1.8\% &  1.6\% &  1.3\% &  1.1\% &  0.8\% &  0.6\% &  0.3\% \\
\bottomrule
\end{tabular}
\caption{Averaged differences between HR values computed on internal temporal windows~($s_2$) located within external ones~($s_1$). Window sizes are given in seconds. Due to $s_1>s_2$, the resulting matrix is lower triangular.}
\label{table:dVal}
\end{table*}
Figure~\ref{fig:DiffBox} illustrates distribution of differences in average HR values computed for pairs of temporal windows $W^{s_1}$ and $W^{s_2}$ for corresponding differences between $s_1$ and $s_2$ sizes.
The average values of differences are presented in Table~\ref{table:dVal}.
\subsection{Results}
The difference between HR values estimated on different temporal windows appears to be up to 10$\%$ and greater which corresponds to 4--10~bpm for normal heart rate at rest (40--100~bpm for healthy adult~\cite{Quer2020}). Also, similar difference is expected to be between HR values estimated by rPPG methods with different processing window sizes. It means that comparison between result of an rPPG method and a reference HR data should be done using the same temporal window sizes; if sizes differ, one should take into account an additional ``error'' caused by the difference.

\paragraph{Acknowledgments.} 
\ifcvprfinal 
The project is supported by the Russian Ministry of Science and Higher Education in the framework of Decree No.~218, project No.~2019-218-11-8185 ``Creating a software complex for human capital management based on neurotechnologies for enterprises of the high-tech sector of the Russian Federation'' (Internal number HD/19-22-NY).
\else
anonymized \\[6\baselineskip]
\fi

\section{Conclusion} 
Several factors and their influences on rPPG methods were considered.
\begin{enumerate}
    \item Rolling shutter effect generates shift of PPG signal which is of the same order of magnitude as ``physiological'' phase shift reported in a number of papers. Thus, it should be properly accounted in studies and interpretations of PPG signal phase shifts.     
    \item In short series of experiments we didn\textquotesingle t observe effect of irregular frame rate in video on distortion of PPG signal spectrum. However, since this irregularity depends on particular recording device and compression method, it could be a subject of separate comprehensive research.
    \item Significant variation of computed ground truth labels of HR is discovered caused by different size of temporal window for HR averaging. In our opinion, size of temporal window both for HR estimation by rPPG method and for ground truth should be explicitly declared when presenting results of evaluation and comparing of different approaches.
\end{enumerate}

{\small
\bibliographystyle{ieeetr}
\bibliography{paper}

\begin{thebibliography}{10}

\bibitem{Prahl2020}
S.~Prahl, ``{Optical Absorption of Hemoglobin.
  [Online]~(\url{https://omlc.org/spectra/hemoglobin/})}.''

\bibitem{Teplov2014}
V.~Teplov, E.~Nippolainen, A.~A. Makarenko, R.~Giniatullin, and A.~A.
  Kamshilin, ``{Ambiguity of mapping the relative phase of blood
  pulsations.},'' {\em Biomedical optics express}, vol.~5, no.~9, pp.~3123--39,
  2014.

\bibitem{Kamshilin2015}
A.~Kamshilin, E.~Nippolainen, I.~Sidorov, P.~Vasilev, N.~Erofeev, N.~Podolian,
  and R.~Romashko, ``A new look at the essence of the imaging
  photoplethysmography,'' {\em Scientific Reports}, no.~5, 2015.

\bibitem{Verkruysse2008}
W.~Verkruysse, L.~O. Svaasand, and J.~S. Nelson, ``{Remote plethysmographic
  imaging using ambient light},'' {\em OPT EXPRESS}, vol.~16, no.~26,
  pp.~21434--21445, 2008.

\bibitem{Wu2000}
T.~Wu, V.~Blazek, and H.~J. Schmitt, ``{Photoplethysmography imaging: a new
  noninvasive and noncontact method for mapping of the dermal perfusion
  changes},'' in {\em Optical Techniques and Instrumentation for the
  Measurement of Blood Composition, Structure, and Dynamics} (A.~V. Priezzhev
  and P.~A. Oberg, eds.), vol.~4163, pp.~62 -- 70, International Society for
  Optics and Photonics, SPIE, 2000.

\bibitem{Wu2003}
T.~Wu, ``Ppgi: New development in noninvasive and contactless diagnosis of
  dermal perfusion using near infrared light,'' {\em Journal of the Society of
  Chinese Physists in Germany}, vol.~7, pp.~17--24, 10 2003.

\bibitem{Wieringa2005}
F.~P. Wieringa, F.~Mastik, and A.~F.~W. {Van Der Steen}, ``{Contactless
  multiple wavelength photoplethysmographic imaging: A first step toward "spO 2
  camera" technology},'' {\em Annals of Biomedical Engineering}, vol.~33,
  no.~8, pp.~1034--1041, 2005.

\bibitem{Haan2013}
G.~{de Haan} and V.~{Jeanne}, ``Robust pulse rate from chrominance-based
  rppg,'' {\em IEEE Transactions on Biomedical Engineering}, vol.~60,
  pp.~2878--2886, Oct 2013.

\bibitem{Wang2017_2}
W.~{Wang}, A.~C. {den Brinker}, S.~{Stuijk}, and G.~{de Haan}, ``Algorithmic
  principles of remote ppg,'' {\em IEEE Transactions on Biomedical
  Engineering}, vol.~64, pp.~1479--1491, July 2017.

\bibitem{Chen2018}
W.~Chen and D.~McDuff, ``Deepphys: Video-based physiological measurement using
  convolutional attention networks,'' in {\em Computer Vision -- ECCV 2018}
  (V.~Ferrari, M.~Hebert, C.~Sminchisescu, and Y.~Weiss, eds.), pp.~356--373,
  Springer International Publishing, 2018.

\bibitem{Yu2019RecoveringRP}
Z.~Yu, X.~Li, and G.~Zhao, ``Recovering remote photoplethysmograph signal from
  facial videos using spatio-temporal convolutional networks,'' {\em ArXiv},
  vol.~abs/1905.02419, 2019.

\bibitem{Sabokrou2020}
M.~Sabokrou, M.~Pourreza, X.~Li, M.~Fathy, and G.~Zhao, ``Deep-hr: Fast heart
  rate estimation from face video under realistic conditions,'' {\em ArXiv},
  2020.

\bibitem{Blackford2017}
E.~B. Blackford and J.~R. Estepp, ``{Measurements of pulse rate using
  long-range imaging photoplethysmography and sunlight illumination
  outdoors},'' in {\em Optical Diagnostics and Sensing XVII: Toward
  Point-of-Care Diagnostics} (G.~L. Coté, ed.), vol.~10072, pp.~122 -- 134,
  International Society for Optics and Photonics, SPIE, 2017.

\bibitem{Finzgar2018}
M.~Finžgar and P.~Podržaj, ``{A wavelet-based decomposition method for a
  robust extraction of pulse rate from video recordings},'' {\em PeerJ},
  vol.~6, 2018.

\bibitem{Poh2010}
M.-Z. Poh, D.~J. McDuff, and R.~W. Picard, ``{Non-contact, automated cardiac
  pulse measurements using video imaging and blind source separation.},'' {\em
  OPT EXPRESS}, vol.~18, no.~10, pp.~10762--10774, 2010.

\bibitem{WangBrinker2018}
W.~Wang, A.~C.~D. Brinker, and G.~de~Haan, ``Single-element remote-ppg,'' {\em
  IEEE Transactions on Biomedical Engineering}, vol.~66, pp.~2032--2043, 2018.

\bibitem{Kooij2019}
K.~M. van~der Kooij and M.~Naber, ``An open-source remote heart rate imaging
  method with practical apparatus and algorithms,'' {\em Behavior research
  methods}, vol.~5, p.~2106–2119, 2019.

\bibitem{Jindal2016}
V.~Jindal, ``Integrating mobile and cloud for ppg signal selection to monitor
  heart rate during intensive physical exercise,'' in {\em Proceedings of the
  International Conference on Mobile Software Engineering and Systems},
  MOBILESoft ’16, (New York, NY, USA), p.~36–37, Association for Computing
  Machinery, 2016.

\bibitem{Olsson2010}
E.~Olsson, ``{Heart rate Variability in Stress-reslated Fatigue, Adolescent
  Anxiety and Depression and its Connection to Lifestyle},'' 2010.

\bibitem{Sun2012}
F.-T. Sun, C.~Kuo, H.-T. Cheng, S.~Buthpitiya, P.~Collins, and M.~Griss,
  ``Activity-aware mental stress detection using physiological sensors,'' in
  {\em Mobile Computing, Applications, and Services} (M.~Gris and G.~Yang,
  eds.), (Berlin, Heidelberg), pp.~211--230, Springer Berlin Heidelberg, 2012.

\bibitem{Mcintyre2016}
S.~Mcintyre, J.~M. Eklund, and C.~Collins, ``{Using Visual Analytics of Heart
  Rate Variation to Aid in Diagnostics},'' in {\em AVI}, pp.~20--27, 2016.

\bibitem{Kamshilin2018_2}
A.~A. Kamshilin, M.~A. Volynsky, O.~Khayrutdinova, D.~Nurkhametova, L.~Babayan,
  A.~V. Amelin, O.~V. Mamontov, and R.~Giniatullin, ``Novel capsaicin-induced
  parameters of microcirculation in migraine patients revealed by imaging
  photoplethysmography,'' {\em J Headache Pain}, vol.~1:43, p.~2106–2119,
  2018.

\bibitem{Rustand2012}
{\AA}.~Rustand, {\em {Ambient-light photoplethysmography}}.
\newblock Doctoral dissertation, master's thesis, Norwegian University of
  Science and Technology, Department of Electronics and Telecommunications,
  2012.

\bibitem{Feng2015a}
L.~Feng, S.~Member, L.-m. Po, and S.~Member, ``{Motion-Resistant Remote Imaging
  Photoplethysmography Based on the Optical Properties of Skin},'' {\em IEEE
  Transactions on Circuits and Systems for Video Technology}, vol.~25, no.~5,
  pp.~879--892, 2015.

\bibitem{Niu2018_1}
X.~Niu, H.~Han, S.~Shan, and X.~Chen, ``Vipl-hr: A multi-modal database for
  pulse estimation from less-constrained face video,'' in {\em Computer Vision
  -- ACCV 2018}, pp.~562--576, 05 2019.

\bibitem{Stricker2014}
R.~{Stricker}, S.~{Müller}, and H.~{Gross}, ``Non-contact video-based pulse
  rate measurement on a mobile service robot,'' in {\em The 23rd IEEE
  International Symposium on Robot and Human Interactive Communication},
  pp.~1056--1062, Aug 2014.

\bibitem{Fletcher2015}
R.~Ribon~Fletcher, D.~Chamberlain, N.~Paggi, and X.~Deng, ``Implementation of
  smart phone video plethysmography and dependence on lighting parameters,'' in
  {\em 2015 37th Annual International Conference of the IEEE Engineering in
  Medicine and Biology Society (EMBC)}, vol.~2015, pp.~3747--3750, 08 2015.

\bibitem{Poh2011}
M.~Z. Poh, D.~J. McDuff, and R.~W. Picard, ``{Advancements in noncontact,
  multiparameter physiological measurements using a webcam},'' {\em IEEE
  Transactions on Biomedical Engineering}, vol.~58, no.~1, pp.~7--11, 2011.

\bibitem{Lewandowska2011}
M.~{Lewandowska}, J.~{Rumiński}, T.~{Kocejko}, and J.~{Nowak}, ``Measuring
  pulse rate with a webcam — a non-contact method for evaluating cardiac
  activity,'' in {\em 2011 Federated Conference on Computer Science and
  Information Systems (FedCSIS)}, pp.~405--410, Sep. 2011.

\bibitem{Tulyakov2016}
S.~{Tulyakov}, X.~{Alameda-Pineda}, E.~{Ricci}, L.~{Yin}, J.~F. {Cohn}, and
  N.~{Sebe}, ``Self-adaptive matrix completion for heart rate estimation from
  face videos under realistic conditions,'' in {\em 2016 IEEE Conference on
  Computer Vision and Pattern Recognition (CVPR)}, pp.~2396--2404, June 2016.

\bibitem{Balakrishnan2013}
G.~Balakrishnan, F.~Durand, and J.~Guttag, ``{Detecting pulse from head motions
  in video},'' in {\em Proceedings of the IEEE Computer Society Conference on
  Computer Vision and Pattern Recognition}, pp.~3430--3437, 2013.

\bibitem{Li2014}
X.~Li, J.~Chen, G.~Zhao, and M.~Pietik, ``{Remote Heart Rate Measurement From
  Face Videos Under Realistic Situations},'' in {\em Proceedings of the IEEE
  Computer Society Conference on Computer Vision and Pattern Recognition},
  2014.

\bibitem{Huang2016}
R.-Y. Huang and L.-R. Dung, ``Measurement of heart rate variability using
  off-the-shelf smart phones,'' {\em BioMedical Engineering OnLine}, vol.~15,
  01 2016.

\bibitem{Haan2014}
G.~de~Haan and A.~van Leest, ``Improved motion robustness of remote-{PPG} by
  using the blood volume pulse signature,'' {\em Physiological Measurement},
  vol.~35, pp.~1913--1926, aug 2014.

\bibitem{Wang2016}
W.~Wang, S.~Stuijk, and G.~D. Haan, ``{A Novel Algorithm for Remote
  Photoplethysmography : Spatial Subspace Rotation},'' {\em IEEE Trans Biomed
  Eng.}, vol.~63, no.~9, pp.~1974--1984, 2016.

\bibitem{Coppetti2017}
T.~Coppetti, A.~Brauchlin, S.~Müggler, A.~Attinger-Toller, C.~Templin,
  F.~Schoenrath, J.~Hellermann, T.~Lüscher, and C.~Wyss, ``Accuracy of
  smartphone apps for heart rate measurement,'' {\em European journal of
  preventive cardiology}, vol.~24, p.~2047487317702044, 05 2017.

\bibitem{Macwan2018}
R.~Macwan, Y.~Benezeth, and A.~Mansouri, ``Remote photoplethysmography with
  constrained ica using periodicity and chrominance constraints,'' {\em
  BioMedical Engineering OnLine}, vol.~17, p.~22, Feb 2018.

\bibitem{Nowara2018}
E.~M. Nowara, T.~K. Marks, H.~Mansour, and A.~Veeraraghavan, ``Sparseppg:
  Towards driver monitoring using camera-based vital signs estimation in
  near-infrared,'' {\em 2018 IEEE/CVF Conference on Computer Vision and Pattern
  Recognition Workshops (CVPRW)}, pp.~1353--135309, 2018.

\bibitem{Kumar2015}
M.~Kumar, A.~Veeraraghavan, and A.~Sabharwal, ``{DistancePPG: Robust
  non-contact vital signs monitoring using a camera},'' {\em Biomedical Optics
  Express}, vol.~6, no.~5, p.~1565, 2015.

\bibitem{Spetlik2018}
R.~Spetlik, J.~Cech, V.~Franc, and J.~Matas, ``Visual heart rate estimation
  with convolutional neural network,'' in {\em Proceedings of British Machine
  Vision Conference}, 2018.

\bibitem{Niu2018_2}
X.~Niu, H.~Han, S.~Shan, and X.~Chen, ``Synrhythm: Learning a deep heart rate
  estimator from general to specific,'' in {\em 2018 24th International
  Conference on Pattern Recognition (ICPR)}, pp.~3580--3585, 2018.

\bibitem{Niu2017}
X.~{Niu}, H.~{Han}, S.~{Shan}, and X.~{Chen}, ``Continuous heart rate
  measurement from face: A robust rppg approach with distribution learning,''
  in {\em 2017 IEEE International Joint Conference on Biometrics (IJCB)},
  pp.~642--650, Oct 2017.

\bibitem{Hsu2014}
Y.~{Hsu}, Y.~{Lin}, and W.~{Hsu}, ``Learning-based heart rate detection from
  remote photoplethysmography features,'' in {\em 2014 IEEE International
  Conference on Acoustics, Speech and Signal Processing (ICASSP)},
  pp.~4433--4437, May 2014.

\bibitem{Estepp2014}
J.~R. {Estepp}, E.~B. {Blackford}, and C.~M. {Meier}, ``Recovering pulse rate
  during motion artifact with a multi-imager array for non-contact imaging
  photoplethysmography,'' in {\em 2014 IEEE International Conference on
  Systems, Man, and Cybernetics (SMC)}, pp.~1462--1469, Oct 2014.

\bibitem{McDuff2014}
D.~J. McDuff, S.~Gontarek, and R.~W. Picard, ``Improvements in remote
  cardiopulmonary measurement using a five band digital camera,'' {\em IEEE
  Transactions on Biomedical Engineering}, vol.~61, pp.~2593--2601, 2014.

\bibitem{Niu2019}
X.~Niu, S.~Shan, H.~Han, and X.~Chen, ``Rhythmnet: End-to-end heart rate
  estimation from face via spatial-temporal representation,'' {\em IEEE
  Transactions on Image Processing}, 10 2019.

\bibitem{yu2019_2}
Z.~Yu, W.~Peng, X.~Li, X.~Hong, and G.~Zhao, ``Remote heart rate measurement
  from highly compressed facial videos: an end-to-end deep learning solution
  with video enhancement,'' in {\em Proceedings of the IEEE International
  Conference on Computer Vision}, pp.~151--160, 2019.

\bibitem{antink2019}
C.~H. Antink, S.~Lyra, M.~Paul, X.~Yu, and S.~Leonhardt, ``A broader look:
  Camera-based vital sign estimation across the spectrum,'' {\em Yearbook of
  medical informatics}, vol.~28, no.~01, pp.~102--114, 2019.

\bibitem{Soleymani2012}
M.~Soleymani, J.~Lichtenauer, T.~Pun, and M.~Pantic, ``{A multimodal database
  for affect recognition and implicit tagging},'' {\em IEEE Transactions on
  Affective Computing}, vol.~3, pp.~42--55, Jan 2012.

\bibitem{Zhang2016}
Z.~{Zhang}, J.~M. {Girard}, Y.~{Wu}, X.~{Zhang}, P.~{Liu}, U.~{Ciftci},
  S.~{Canavan}, M.~{Reale}, A.~{Horowitz}, H.~{Yang}, J.~F. {Cohn}, Q.~{Ji},
  and L.~{Yin}, ``Multimodal spontaneous emotion corpus for human behavior
  analysis,'' in {\em 2016 IEEE Conference on Computer Vision and Pattern
  Recognition (CVPR)}, pp.~3438--3446, 2016.

\bibitem{mcduff2019}
D.~McDuff and E.~Blackford, ``iphys: An open non-contact imaging-based
  physiological measurement toolbox,'' in {\em 2019 41st Annual International
  Conference of the IEEE Engineering in Medicine and Biology Society (EMBC)},
  pp.~6521--6524, IEEE, 2019.

\bibitem{pilz2019}
C.~Pilz, ``On the vector space in photoplethysmography imaging,'' in {\em
  Proceedings of the IEEE International Conference on Computer Vision
  Workshops}, 2019.

\bibitem{Li2020}
X.~Li, H.~Han, H.~Lu, X.~Niu, Z.~Yu, A.~Dantcheva, G.~Zhao, and S.~Shan, ``{The
  1st Challenge on Remote Physiological Signal Sensing (RePSS)},'' {\em arXiv},
  2020.

\bibitem{mcduff2018}
D.~McDuff, ``Deep super resolution for recovering physiological information
  from videos,'' in {\em Proceedings of the IEEE Conference on Computer Vision
  and Pattern Recognition Workshops}, pp.~1367--1374, 2018.

\bibitem{zhao2018}
C.~Zhao, C.-L. Lin, W.~Chen, and Z.~Li, ``A novel framework for remote
  photoplethysmography pulse extraction on compressed videos,'' in {\em
  Proceedings of the IEEE Conference on Computer Vision and Pattern Recognition
  Workshops}, pp.~1299--1308, 2018.

\bibitem{moco2018_1}
A.~{Moço}, S.~{Stuijk}, M.~{van Gastel}, and G.~{de Haan}, ``Impairing factors
  in remote-ppg pulse transit time measurements on the face,'' in {\em 2018
  IEEE/CVF Conference on Computer Vision and Pattern Recognition Workshops
  (CVPRW)}, pp.~1439--14398, June 2018.

\bibitem{Oth2013}
L.~{Oth}, P.~{Furgale}, L.~{Kneip}, and R.~{Siegwart}, ``Rolling shutter camera
  calibration,'' in {\em 2013 IEEE Conference on Computer Vision and Pattern
  Recognition}, pp.~1360--1367, June 2013.

\bibitem{Lao2018}
Y.~Lao and O.~Ait-Aider, ``A robust method for strong rolling shutter effects
  correction using lines with automatic feature selection,'' in {\em 2018
  IEEE/CVF Conference on Computer Vision and Pattern Recognition (CVPR)}, (Los
  Alamitos, CA, USA), pp.~4795--4803, IEEE Computer Society, jun 2018.

\bibitem{GnuRadio}
``{GNU Radio - The free \& open software radio ecosystem.
  [Online]~(\url{http://www.gnuradio.org})}.''

\bibitem{Iyengar1996}
N.~Iyengar, C.-K. Peng, R.~Morin, A.~Goldberger, and L.~Lipsitz, ``Age-related
  alterations in the fractal scaling of cardiac interbeat interval dynamics,''
  {\em The American journal of physiology}, vol.~271, pp.~R1078--84, 11 1996.

\bibitem{Goldberger2000}
A.~Goldberger, L.~Amaral, L.~Glass, J.~Hausdorff, P.~Ivanov, R.~Mark,
  J.~Mietus, G.~Moody, C.-K. Peng, and H.~Stanley, ``Physiobank, physiotoolkit,
  and physionet : Components of a new research resource for complex physiologic
  signals,'' {\em Circulation}, vol.~101, pp.~E215--20, 07 2000.

\bibitem{Quer2020}
G.~Quer, P.~Gouda, M.~Galarnyk, E.~Topol, and S.~Steinhubl, ``Inter- and
  intraindividual variability in daily resting heart rate and its associations
  with age, sex, sleep, bmi, and time of year: Retrospective, longitudinal
  cohort study of 92,457 adults,'' {\em PLOS ONE}, vol.~15, p.~e0227709, 02
  2020.

\end{thebibliography}
}

\end{document}